\documentclass[conference]{IEEEtran}
\IEEEoverridecommandlockouts
\usepackage{cite}
\usepackage{amsmath,amssymb,amsfonts}
\usepackage{graphicx}
\usepackage{textcomp}
\usepackage{xcolor}
\usepackage{tikz}
\usepackage{pgfplots}
\usepackage{booktabs}
\usepackage{multirow}
\usepackage{array}
\usepackage{subcaption}
\usepackage{hyperref}
\usepackage{hyphenat}

\usetikzlibrary{arrows.meta, positioning, shapes.geometric, fit, calc, backgrounds}
\pgfplotsset{compat=1.18}

\def\BibTeX{{\rm B\kern-.05em{\sc i\kern-.025em b}\kern-.08em
    T\kern-.1667em\lower.7ex\hbox{E}\kern-.125emX}}
\begin{document}

\makeatletter
\def\ps@IEEEtitlepagestyle{%
  \def\@oddfoot{\mycopyrightnotice}%
  \def\@evenfoot{}%
}
\def\mycopyrightnotice{%
  {\footnotesize
    \parbox[b]{\textwidth}{%
      \copyright~2026 IEEE. Personal use of this material is permitted.
      Permission from IEEE must be obtained for all other uses, in any current
      or future media, including reprinting/republishing this material for
      advertising or promotional purposes, creating new collective works, for
      resale or redistribution to servers or lists, or reuse of any copyrighted
      component of this work in other works.%
    }%
  }%
  \gdef\mycopyrightnotice{}
}
\makeatother

\title{Real-Time Video Anomaly Detection Using YOLO Pose Estimation and CLIP-Based Semantic Scoring}

\author{\IEEEauthorblockN{Vanodhya G. Warnasooriya\textsuperscript{1},
Amir Hajian\textsuperscript{1},
Watchara Ruangsang\textsuperscript{2},
and Supavadee Aramvith\textsuperscript{1}}
\IEEEauthorblockA{\textsuperscript{1}\textit{Dept. of Electrical Engineering, Faculty of Engineering, Chulalongkorn University}, Bangkok 10330, Thailand}
\IEEEauthorblockA{\textsuperscript{2}\textit{Media Technology Program, King Mongkut's Univ. of Technology Thonburi}, Bangkok 10150, Thailand}
\IEEEauthorblockA{\{vanowarna, amirhajian85\}@gmail.com, watchara.ruan@kmutt.ac.th, supavadee.a@chula.ac.th}
}

\maketitle

\begin{abstract}
We propose a lightweight two-stage framework for real-time video anomaly detection. The first stage employs YOLO~v11n-pose to detect persons and extract seventeen skeletal keypoints in a single forward pass. The second stage encodes each cropped person region through CLIP~ViT-B/32 and computes cosine similarity against predefined textual descriptions of anomalous behaviors. This architecture eliminates the need for optical flow, standalone pose estimators, and density-based scoring modules. Experiments on CUHK~Avenue, ShanghaiTech~Campus, and a custom indoor dataset collected at Chulalongkorn~University demonstrate an end-to-end throughput of approximately 51~FPS on an NVIDIA Titan~XP GPU, a 3.36$\times$ speedup over the multi-feature baseline, while maintaining frame-level AUROC values of 89.26\%, 70.26\%, and 84.13\%, respectively.
\end{abstract}


\section{Introduction}
\label{sec:intro}

The widespread deployment of closed-circuit television (CCTV) across urban and institutional environments has driven growing demand for intelligent video analytics~\cite{b1}. Human operators monitoring multiple feeds are susceptible to attention fatigue, with effectiveness declining sharply within minutes~\cite{b2}, motivating automated systems for detecting falls, altercations, and abnormal postures.

Prior work on video anomaly detection spans recon\-struction-based~\cite{b3,b4}, prediction-based~\cite{b5}, and attribute-based~\cite{b6} paradigms. Reiss and Hoshen~\cite{b6} achieved strong accuracy by combining AlphaPose, FlowNet2, and CLIP features with GMM/kNN scoring, but the multi-component pipeline limits throughput to approximately 15~FPS. Two recent developments enable a simpler design: CLIP~\cite{b7} provides zero-shot anomaly classification via natural-language prompts without task-specific anomaly labels, and YOLO~v11n-pose~\cite{b8} performs person detection with seventeen\hyp{}keypoint regression in a single forward pass. Although VadCLIP~\cite{b9} also adapts CLIP for video anomaly detection, it relies on learnable prompt tokens trained with weak supervision and is designed for offline analysis rather than real-time deployment.

This work targets person-level anomalies (specifically falling, lying on the floor, fighting, and sitting on the floor) in fixed indoor CCTV environments. The system is designed for deployment on a single desktop-class GPU (NVIDIA Titan~XP) to sustain real-time throughput above 30~FPS on standard surveillance video feeds.

Building on these foundations, we present a two-stage framework in which YOLO~v11n-pose handles detection and CLIP~ViT-B/32 handles semantic anomaly scoring. Our contributions are as follows:

\begin{itemize}
    \item A streamlined pipeline that replaces multi-feature density estimation with direct CLIP semantic scoring, yielding a $3.36\times$ throughput gain over the prior baseline while maintaining comparable accuracy.
    \item A text-prompt-based classification mechanism that leverages CLIP's zero-shot capability to identify and categorize multiple anomaly types without requiring labeled anomaly training data.
    \item Evaluation on three surveillance benchmarks and deployment on live CCTV feeds at Chulalongkorn~University, confirming stable real-time operation.
\end{itemize}

\section{Related Work}
\label{sec:related}

Video anomaly detection methods broadly follow recon\-struction-based or prediction-based paradigms. Hasan~et~al.~\cite{b3} trained convolutional autoencoders on normal video sequences and flagged high reconstruction error as anomalous. Ganokratanaa~et~al.~\cite{b10} proposed a deep spatio\-temporal translation network (DSTN) that fuses background-removed frames with optical flow for pixel-level localization. Morais~et~al.~\cite{b11} introduced MPED-RNN, which learns normal motion regularity from skeleton trajectories using a multi-timescale recurrent architecture and detects anomalies as deviations from predicted joint positions. Reiss and Hoshen~\cite{b6} combined pose (AlphaPose), velocity (FlowNet2), and deep features (CLIP) with GMM/kNN density estimation, achieving strong accuracy but at a throughput of only 15~FPS due to the multi-component pipeline.

Recent vision-language models offer a promising alternative. CLIP~\cite{b7} maps images and text into a shared embedding space through contrastive pre-training, enabling zero-shot classification via natural-language prompts. Wu~et~al.~\cite{b9} developed VadCLIP, which adapts CLIP for weakly supervised video anomaly detection using learnable prompts. Moriyama~et~al.~\cite{b12} proposed a language-guided decision override that adapts CLIP for retraining-free anomaly detection. However, these methods typically operate offline and are not designed for real-time deployment. Our work shares the use of vision-language models with these approaches~\cite{b6,b9,b12}; the contribution lies in system integration and architectural simplification, demonstrating that direct CLIP cosine similarity can replace multi-component density estimation for real-time person-level anomaly detection rather than introducing a fundamentally new scoring mechanism.

On the detection side, YOLO~v11n-pose~\cite{b8} performs person detection and seventeen-keypoint regression in a single forward pass using Cross-Stage Partial Networks with compact C3K2 blocks, providing a favorable accuracy-speed balance for real-time surveillance.

\section{Proposed Method}
\label{sec:method}

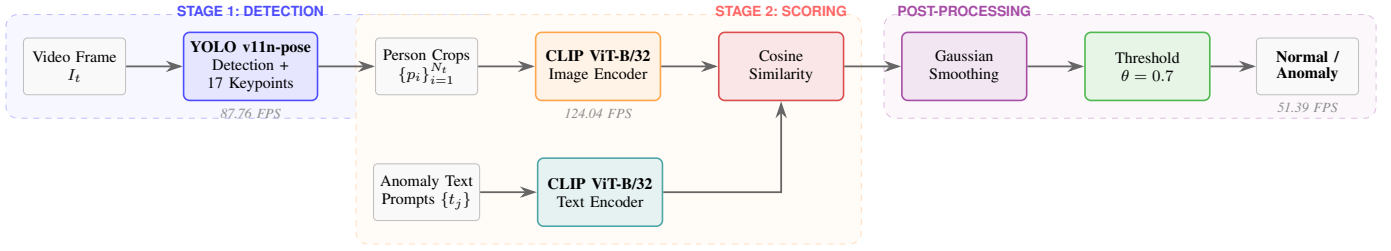
\begin{figure*}[t]
\centering
\resizebox{\textwidth}{!}{%
\begin{tikzpicture}[
    node distance=1.0cm,
    block/.style={rectangle, draw=#1!70, fill=#1!10, minimum height=1.2cm, minimum width=2.2cm, align=center, font=\footnotesize, rounded corners=3pt, line width=0.8pt},
    ioblock/.style={rectangle, draw=gray!50, fill=gray!5, minimum height=1.0cm, minimum width=1.8cm, align=center, font=\footnotesize, rounded corners=2pt, line width=0.5pt},
    arr/.style={-{Stealth[length=3mm, width=2mm]}, line width=0.9pt, color=black!60},
]

\node[ioblock] (input) {Video Frame\\$I_t$};
\node[block=blue, right=1.0cm of input] (yolo) {\textbf{YOLO v11n-pose}\\Detection +\\17 Keypoints};
\node[ioblock, right=1.0cm of yolo] (crops) {Person Crops\\$\{p_i\}_{i=1}^{N_t}$};
\node[block=orange, right=1.0cm of crops] (clipimg) {\textbf{CLIP ViT-B/32}\\Image Encoder};
\node[block=red!80!black, right=1.0cm of clipimg] (cosine) {Cosine\\Similarity};
\node[block=violet, right=1.0cm of cosine] (smooth) {Gaussian\\Smoothing};
\node[block=green!60!black, right=1.0cm of smooth] (thresh) {Threshold\\$\theta = 0.7$};
\node[ioblock, right=0.8cm of thresh] (out) {\textbf{Normal /}\\[-1pt]\textbf{Anomaly}};

\node[ioblock, below=1.2cm of crops] (prompts) {Anomaly Text\\Prompts $\{t_j\}$};
\node[block=teal, right=1.0cm of prompts] (cliptxt) {\textbf{CLIP ViT-B/32}\\Text Encoder};

\draw[arr] (input) -- (yolo);
\draw[arr] (yolo) -- (crops);
\draw[arr] (crops) -- (clipimg);
\draw[arr] (clipimg) -- (cosine);
\draw[arr] (cosine) -- (smooth);
\draw[arr] (smooth) -- (thresh);
\draw[arr] (thresh) -- (out);

\draw[arr] (prompts) -- (cliptxt);
\draw[arr] (cliptxt) -| (cosine);

\node[above=0.15cm of yolo, font=\scriptsize\bfseries\sffamily, text=blue!70] {STAGE 1: DETECTION};
\node[above=0.15cm of cosine, font=\scriptsize\bfseries\sffamily, text=red!60] {STAGE 2: SCORING};
\node[above=0.15cm of smooth, font=\scriptsize\bfseries\sffamily, text=violet!70] {POST-PROCESSING};

\node[below=0.0cm of yolo.south, anchor=north, font=\scriptsize\itshape, text=black!45] {87.76 FPS};
\node[below=0.0cm of clipimg.south, anchor=north, font=\scriptsize\itshape, text=black!45] {124.04 FPS};
\node[below=0.0cm of out.south, anchor=north, font=\scriptsize\itshape, text=black!45] {51.39 FPS};

\begin{scope}[on background layer]
    \node[draw=blue!25, dashed, fill=blue!3, rounded corners=6pt, inner sep=0.3cm, fit=(input)(yolo)(crops)] {};
    \node[draw=orange!25, dashed, fill=orange!3, rounded corners=6pt, inner sep=0.3cm, fit=(clipimg)(cosine)(cliptxt)(prompts)] {};
    \node[draw=violet!25, dashed, fill=violet!3, rounded corners=6pt, inner sep=0.3cm, fit=(smooth)(thresh)(out)] {};
\end{scope}

\end{tikzpicture}%
}
\caption{Overview of the proposed anomaly detection pipeline. Stage~1 employs YOLO~v11n-pose for person detection and skeletal keypoint extraction. Stage~2 encodes each cropped person region via the CLIP image encoder and computes cosine similarity against textual anomaly prompts from the CLIP text encoder. The frame-level anomaly score is then temporally smoothed and compared against a threshold for final classification.}
\label{fig:pipeline}
\end{figure*}

The proposed framework processes each video frame through two sequential stages: person detection with pose estimation, followed by CLIP-based semantic anomaly scoring. Fig.~\ref{fig:pipeline} illustrates the complete pipeline.

\subsection{Stage 1: Person Detection and Pose Estimation}

Given an input frame $I_t \in \mathbb{R}^{H \times W \times 3}$ at time step $t$, the first stage applies YOLO~v11n-pose to simultaneously detect all persons and regress their skeletal poses. For each detected person $i$, the model outputs a bounding box $b_i = (x_i, y_i, w_i, h_i)$ with confidence $c_i$, alongside seventeen keypoints $\mathbf{k}_i = \{(x_j^k, y_j^k, v_j^k)\}_{j=1}^{17}$ following the COCO topology (nose, eyes, ears, shoulders, elbows, wrists, hips, knees, and ankles). The keypoint visibility flags, $v_j^k$, indicate whether each joint is visible, partially occluded, or not detected. Fig.~\ref{fig:qualitative}(a) shows an example with an overlaid keypoint skeleton and a bounding box.

Each detected person is cropped from the original frame using the bounding box coordinates with a padding margin $\delta = 10$~pixels to preserve contextual information:
\begin{equation}
p_i = I_t[y_i \!-\! \delta : y_i \!+\! h_i \!+\! \delta, \; x_i \!-\! \delta : x_i \!+\! w_i \!+\! \delta]
\label{eq:crop}
\end{equation}
producing a set of person crops $\{p_i\}_{i=1}^{N_t}$, where $N_t$ denotes the number of detections at time $t$. The padding value $\delta = 10$~pixels was selected empirically on CU~Indoor~Anomaly frames (native resolution $640 \times 480$), where it corresponds to approximately 1.6\% of the frame width and provides sufficient surrounding context for the downstream CLIP encoder without introducing excessive background clutter.

\subsection{Stage 2: CLIP-Based Semantic Anomaly Scoring}

Each person crop $p_i$ is resized to $224 \times 224$ pixels and encoded by the CLIP~ViT-B/32 visual encoder into a normalized embedding $\mathbf{f}_i^{\text{img}} \in \mathbb{R}^{512}$. A fixed set of $M = 4$ textual prompts describing anomalous behaviors (``A person lying on the floor'', ``A person falling down'', ``A person fighting with another person'', ``A person sitting on the floor'') is similarly encoded into textual embeddings $\mathbf{f}_j^{\text{txt}} \in \mathbb{R}^{512}$, computed once at initialization and cached.

The per-person anomaly score is the maximum cosine similarity across all prompts:
\begin{equation}
s_i^t = \max_{j \in \{1,\ldots,M\}} \; \mathbf{f}_i^{\text{img}} \cdot \mathbf{f}_j^{\text{txt}}
\label{eq:score_person}
\end{equation}
and the frame-level score aggregates across all detected persons:
\begin{equation}
S_t = \max_{i \in \{1,\ldots,N_t\}} \; s_i^t
\label{eq:score_frame}
\end{equation}

\subsection{Temporal Smoothing and Classification}

Raw scores are smoothed with a one-dimensional Gaussian kernel ($\sigma = 5$, half-window $w = 15$) to suppress transient fluctuations. The smoothed score $\hat{S}_t$ is compared against a decision threshold $\theta$; a frame is classified as anomalous when $\hat{S}_t \geq \theta$. The optimal threshold $\theta^* = 0.7$ is selected by sweeping $[0.1, 0.9]$ on a validation partition to maximize the AUROC.

\section{Experimental Setup}
\label{sec:setup}

\subsection{Datasets}

We evaluate on three datasets: \textbf{CUHK Avenue}~\cite{b13} (37 videos, 30,652 frames from a fixed university pathway camera), \textbf{ShanghaiTech Campus}~\cite{b14} (437 videos, 317,398 frames from 13 diverse campus scenes), and \textbf{CU Indoor Anomaly}, a custom dataset of 40 videos (17,798 frames; 11,841 normal, 5,957 anomalous) collected from CCTV cameras in four buildings at Chulalongkorn~University, with frame-level annotations for four anomaly categories (falling, lying, sitting on floor, fighting). The dataset is partitioned into 33 training and 7 test videos.

\subsection{Implementation Details}

YOLO~v11n-pose is initialized from COCO-pretrained weights and fine-tuned on 2,864 training images (augmented from 1,154 manually labeled frames) from CU~Indoor~Anomaly for 500 epochs ($640 \times 640$ input, batch~16, SGD with lr\,=\,0.01) on an NVIDIA Titan~XP GPU\@. For CUHK~Avenue and ShanghaiTech, the same fine-tuned model is used; COCO pretraining provides sufficient general\-ization for person detection across diverse scenes, and the CU~Indoor fine-tuning primarily improves bounding-box precision for indoor poses. CLIP~ViT-B/32 is used with pre-trained weights without fine-tuning; input crops are resized to $224 \times 224$ following the standard CLIP pre\-processing. We note that the term ``zero-shot'' in this work refers specifically to the CLIP-based anomaly scoring component, which requires no labeled anomaly examples; the YOLO detector is supervised for person localization.

\subsection{Evaluation Protocol}

We adopt the frame-level area under the receiver operating characteristic curve (AUROC) as the primary evaluation metric, following established practice in the anomaly detection literature~\cite{b6,b13,b14}. AUROC quantifies the probability that a randomly sampled anomalous frame receives a higher anomaly score than a randomly sampled normal frame. For YOLO~v11n-pose, we additionally report precision, recall, and mean average precision at IoU thresholds of 0.5 (mAP@.5) and 0.5:0.95 (mAP@.5:.95), evaluated on the CU~Indoor~Anomaly test partition. Throughput is measured as end-to-end frames per second on a single NVIDIA Titan~XP GPU, inclusive of all preprocessing, inference, and post-processing stages.

\section{Results and Analysis}
\label{sec:results}

\subsection{Anomaly Detection Performance}

Table~\ref{tab:auroc} reports frame-level AUROC results. On CUHK~Avenue, the proposed framework achieves 89.26\%, matching the multi-feature baseline employing FlowNet2, AlphaPose, and GMM/kNN scoring. On the ShanghaiTech~Campus, the score of 70.26\% reflects the challenge posed by highly diverse scenes and anomaly types (cycling, skateboarding, vehicle intrusion) that fall outside the coverage of the four fixed text prompts. On CU~Indoor~Anomaly, the method reaches 84.13\%, surpassing the baseline by roughly two percentage points. We attribute this gain to the strong semantic alignment between CLIP's text-image similarity and the specific anomaly categories (falling, lying, fighting, sitting) present in this dataset. Notably, the proposed method outperforms DSTN~\cite{b10} on Avenue by 2.86 percentage points while operating at substantially higher throughput. To clarify the Table~\ref{tab:auroc} footnotes: the multi-feature baseline is our reimplementation of~\cite{b6} with a YOLO-based detector and scores anomalies via GMM/kNN density estimation over concatenated pose, velocity, and deep features, whereas the proposed method scores via direct CLIP text-image cosine similarity. Both pipelines share the same YOLO detector, so Avenue and ShanghaiTech values are identical; the AUROC difference on CU~Indoor isolates the effect of the scoring mechanism.

\begin{table}[t]
\caption{Frame-Level AUROC (\%) Comparison with Existing Methods}
\label{tab:auroc}
\centering
\renewcommand{\arraystretch}{1.15}
\begin{tabular}{lccc}
\toprule
\textbf{Method} & \textbf{Avenue} & \textbf{SHTech} & \textbf{CU Indoor} \\
\midrule
Conv-AE~\cite{b3}              & 80.00 & 60.85 & -- \\
MPED-RNN~\cite{b11}            & 83.50 & 73.40 & -- \\
DSTN~\cite{b10}                 & 86.40 & 73.10 & -- \\
Attr-Based~\cite{b6}           & 86.00 & 71.10 & -- \\
Multi-feat.~baseline$^\dagger$ & 89.26 & 70.26 & 82.12 \\
\midrule
\textbf{Proposed}              & \textbf{89.26}$^\ddagger$ & \textbf{70.26}$^\ddagger$ & \textbf{84.13} \\
\bottomrule
\multicolumn{4}{l}{\scriptsize $^\dagger$Our reimplementation of~\cite{b6} with YOLO-based detection.}\\
\multicolumn{4}{l}{\scriptsize $^\ddagger$Shares the same YOLO detector and evaluation as $^\dagger$;}\\
\multicolumn{4}{l}{\scriptsize only the scoring stage differs (CLIP vs.\ density estimation).}
\end{tabular}
\end{table}

\subsection{Qualitative Detection Results}

\begin{figure}[t]
\centering
\begin{subfigure}[b]{\columnwidth}
    \includegraphics[width=\linewidth]{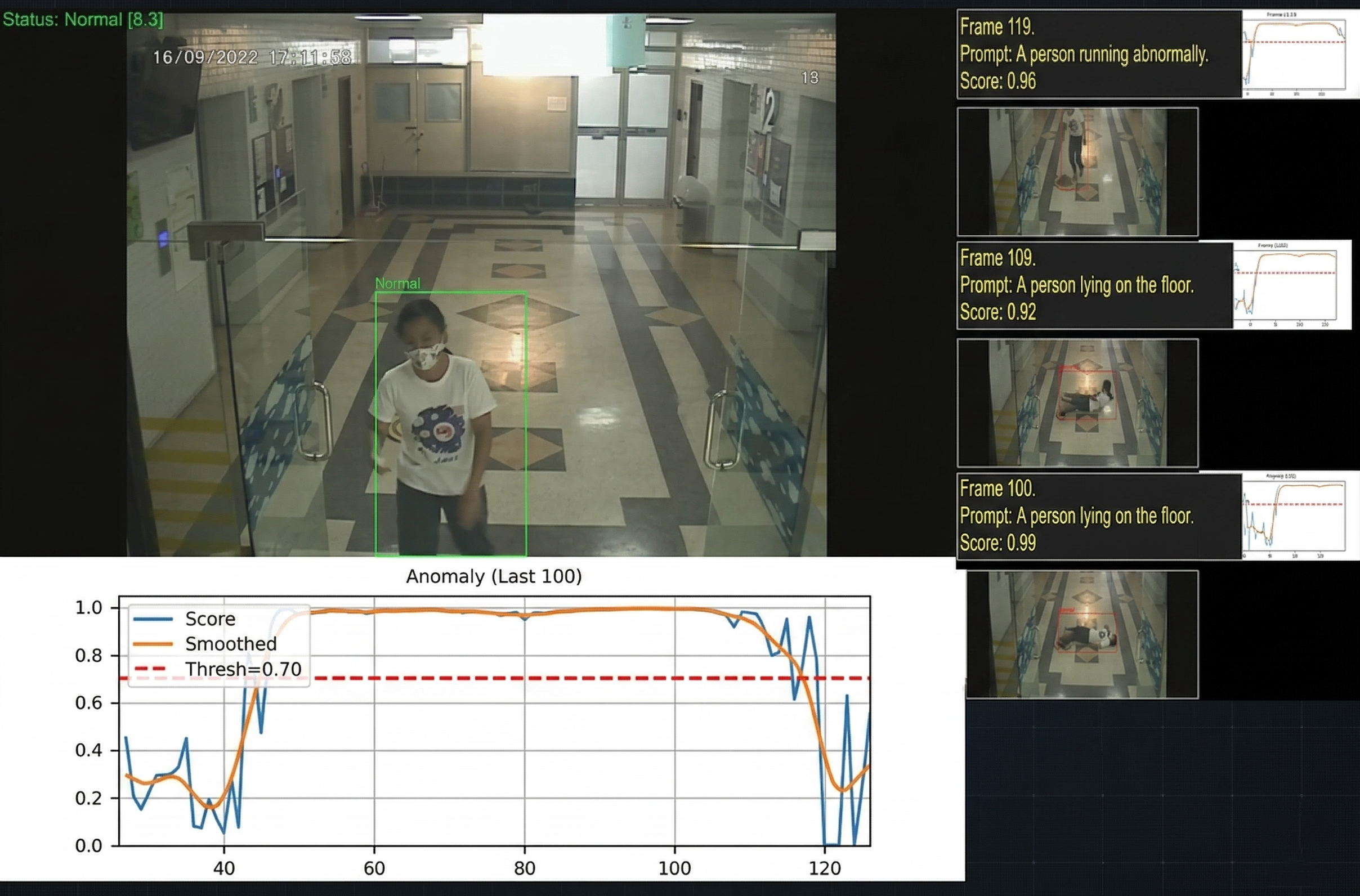}
    \caption{GUI output during normal behavior (walking)}
    \label{fig:qual_a}
\end{subfigure}
\\[3pt]
\begin{subfigure}[b]{\columnwidth}
    \includegraphics[width=\linewidth]{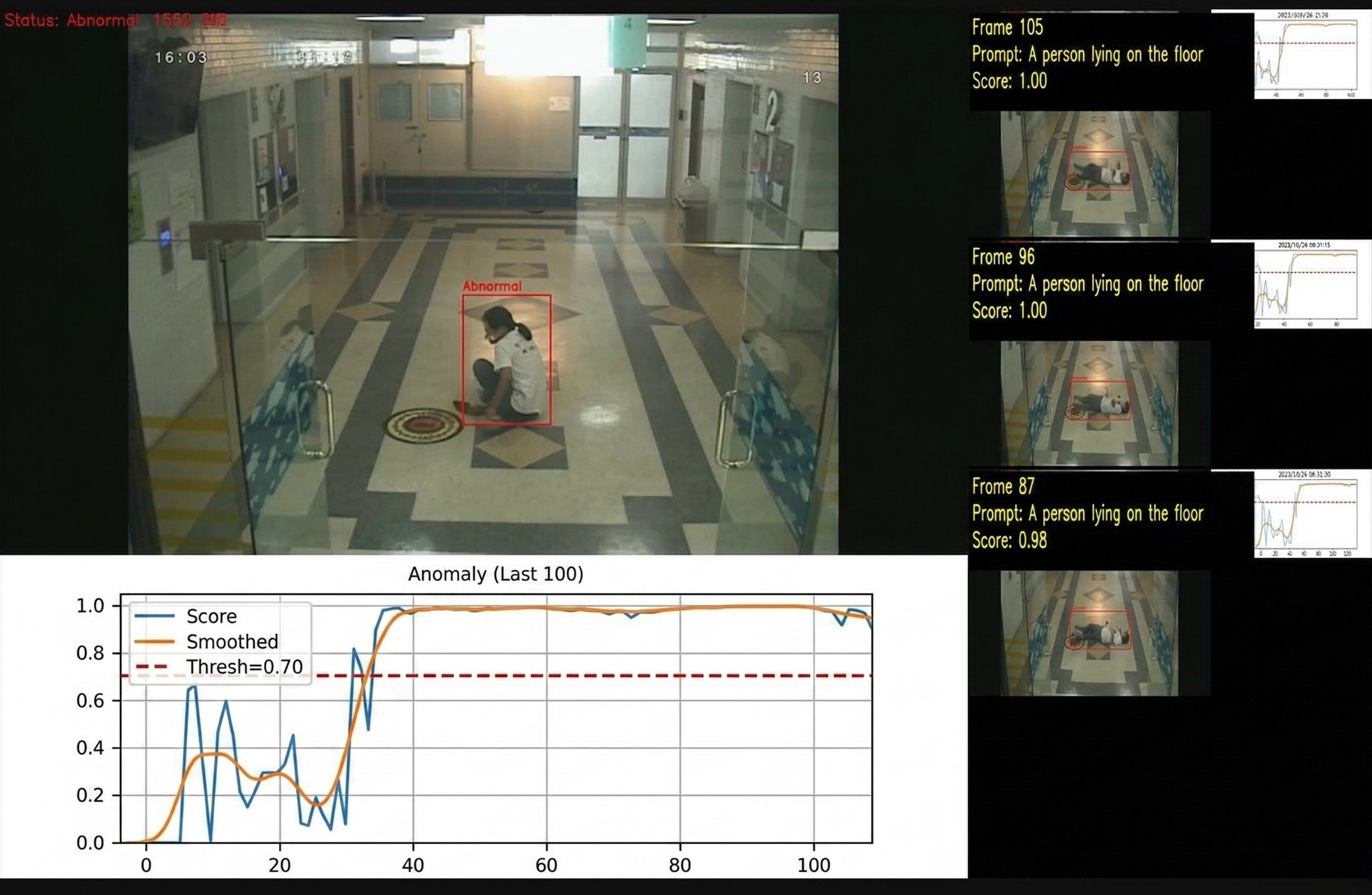}
    \caption{GUI output during detected anomaly (lying on floor)}
    \label{fig:qual_b}
\end{subfigure}
\\[3pt]
\begin{subfigure}[b]{0.48\columnwidth}
    \includegraphics[width=\linewidth]{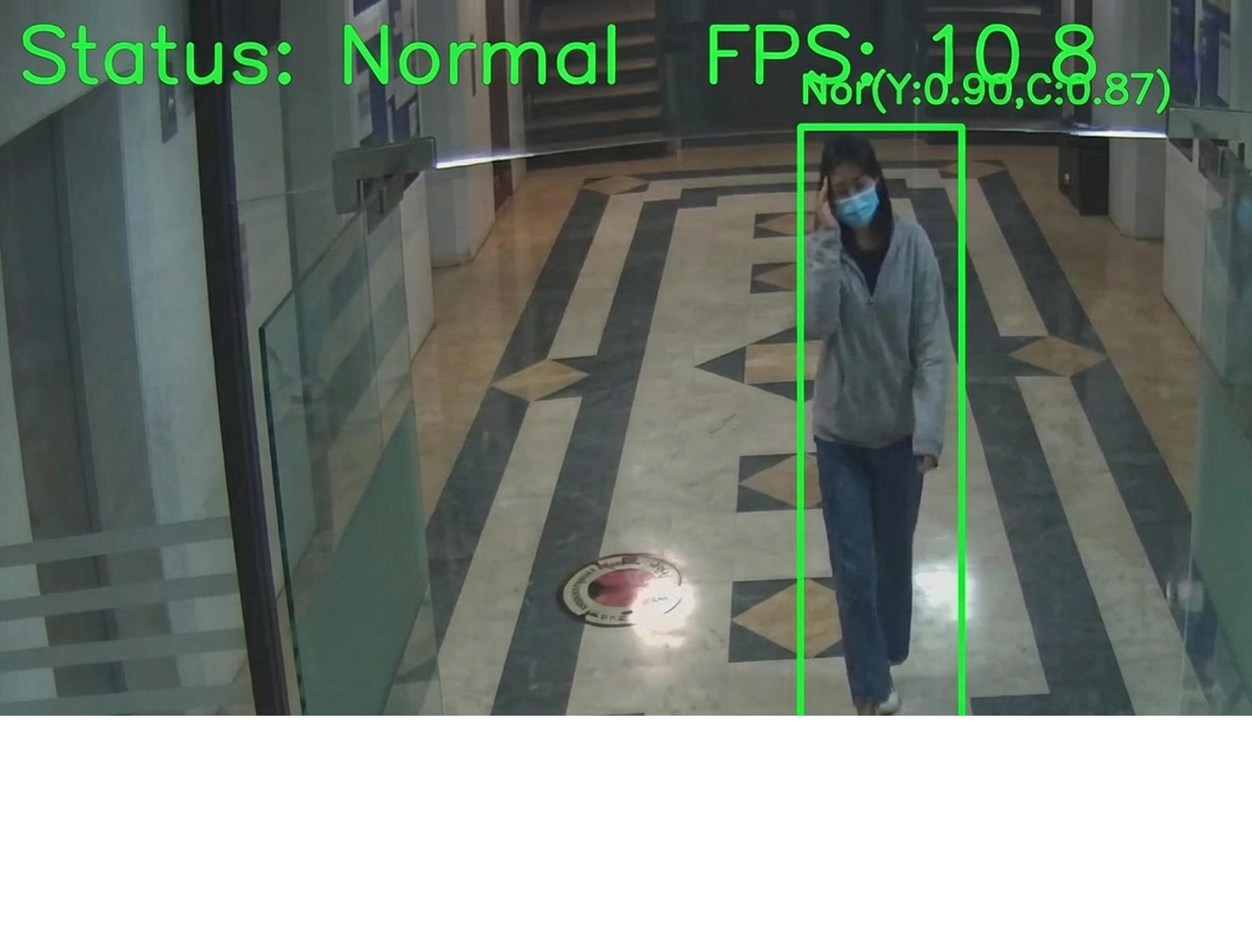}
    \caption{Normal: bounding box}
    \label{fig:qual_c}
\end{subfigure}
\hfill
\begin{subfigure}[b]{0.48\columnwidth}
    \includegraphics[width=\linewidth]{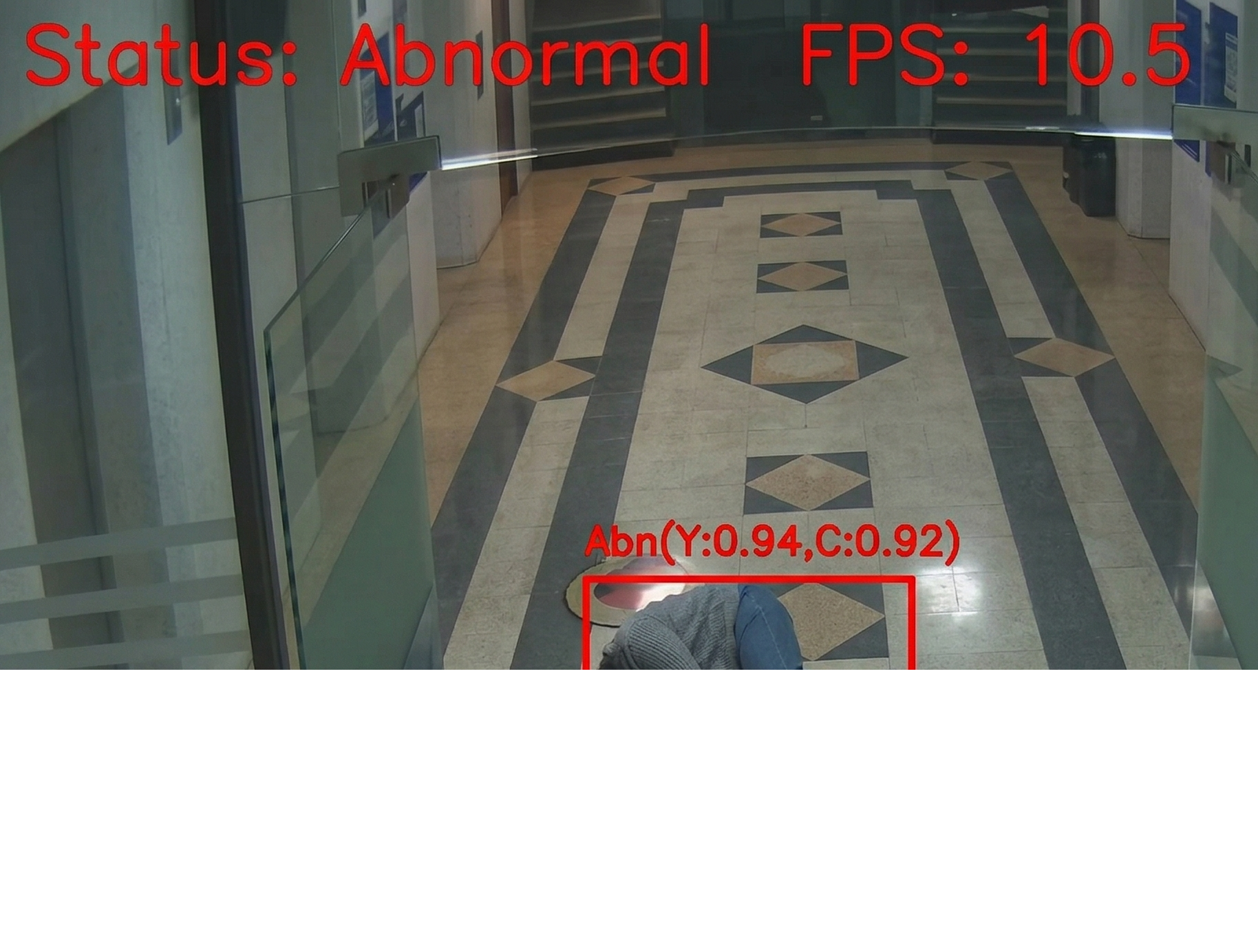}
    \caption{Abnormal: falling}
    \label{fig:qual_d}
\end{subfigure}
\caption{Real-time detection results on the CU~Indoor~Anomaly dataset. (a)--(b)~Full system GUI showing live video feed, real-time anomaly score graph, and frame history panel for normal and abnormal scenarios. (c)--(d)~Close-up views showing green bounding boxes for normal behavior and red bounding boxes with anomaly scores for detected anomalies.}
\label{fig:qualitative}
\end{figure}

Fig.~\ref{fig:qualitative} shows representative outputs. Normal persons are enclosed in green bounding boxes (a,~c), while detected anomalies are enclosed in red boxes with YOLO confidence and CLIP scores (b,~d).

\subsection{YOLO Detection Performance}

Table~\ref{tab:yolo} reports the fine-tuned YOLO~v11n-pose detection metrics on the CU~Indoor~Anomaly test set, evaluated across 237 images containing 360 annotated person instances. The model achieves 91\% precision and 97\% recall overall, with mAP@.5 of 92\%. Normal poses are detected with higher precision (95\%) than abnormal ones (86\%), which is expected given the greater variability and less stereotyped configurations associated with anomalous body positions such as lying and sitting on the floor. The high recall across both categories (98\% normal, 96\% abnormal) confirms that the detector rarely misses persons in the scene, which is critical for the downstream CLIP scoring stage. The mAP@.5:.95 gap between the normal (71\%) and abnormal (63\%) classes suggests that bounding-box localization is less precise for abnormal poses, though this has minimal impact on CLIP scoring because the crops include contextual padding.

\begin{table}[t]
\caption{YOLO v11n-pose Detection Performance on CU Indoor Anomaly}
\label{tab:yolo}
\centering
\renewcommand{\arraystretch}{1.15}
\begin{tabular}{lcccc}
\toprule
\textbf{Category} & \textbf{Prec.} & \textbf{Rec.} & \textbf{mAP@.5} & \textbf{mAP@.5:.95} \\
\midrule
All      & 0.91 & 0.97 & 0.92 & 0.67 \\
Normal   & 0.95 & 0.98 & 0.96 & 0.71 \\
Abnormal & 0.86 & 0.96 & 0.88 & 0.63 \\
\bottomrule
\end{tabular}
\end{table}

\subsection{Processing Throughput}

Table~\ref{tab:speed} compares per-component and end-to-end throughput between the baseline and proposed pipelines. The \textit{baseline} employs Mask~R-CNN for detection, FlowNet2 for optical flow, AlphaPose for skeleton estimation, and GMM/kNN for density-based scoring. The \textit{proposed} method removes all intermediate modules and relies exclusively on YOLO~v11n-pose and CLIP, achieving an end-to-end throughput of 51.39~FPS, a $3.36\times$ speedup over the baseline. The bottleneck shifts from object detection (23.41~FPS in the baseline) to YOLO~v11n-pose (87.76~FPS), with CLIP scoring running comfortably at 124.04~FPS. Note that the baseline density estimation speed (2578.80~FPS) reflects the lightweight nature of GMM/kNN scoring on pre-extracted feature vectors. The gap between per-component rates and the measured end-to-end throughput arises from overhead not captured by isolated benchmarks: video frame decoding from the CCTV stream, CPU-to-GPU data transfer for each person crop, variable CLIP batch sizes depending on the number of persons detected per frame, and Python/OpenCV rendering of bounding-box overlays. These overheads are inherent to a live deployment pipeline.

\begin{table}[t]
\caption{Processing Throughput Comparison (Frames Per Second)}
\label{tab:speed}
\centering
\renewcommand{\arraystretch}{1.15}
\begin{tabular}{lcc}
\toprule
\textbf{Component} & \textbf{Baseline} & \textbf{Proposed} \\
\midrule
Object detection     & 23.41   & 87.76 \\
Deep features (CLIP) & 45.11   & -- \\
Density estimation   & 2578.80 & -- \\
CLIP scoring         & --      & 124.04 \\
\midrule
\textbf{End-to-end}  & \textbf{15.32} & \textbf{51.39} \\
\bottomrule
\end{tabular}
\end{table}

\section{Discussion}
\label{sec:discussion}

The principal strength of the proposed framework is its architectural simplicity. By routing person crops directly through CLIP for text-image similarity, the pipeline eliminates optical flow extraction (FlowNet2), standalone skeleton estimation (AlphaPose), and iterative density model fitting (GMM/kNN), reducing the system to two neural network models. This accelerates throughput from 15.32 to 51.39~FPS while requiring only a single GPU and two model checkpoints for deployment.

The text-prompt mechanism provides practical flexibility: adapting the system to a new surveillance environment requires only updating the prompt set, with no retraining. CLIP's zero-shot capability further enables generalization to anomaly categories absent from the training data, provided they can be described in natural language. The system was deployed on five live CCTV feeds in the elevator area on the 12th floor of the Charoen Wisawakam Building at Chulalongkorn~University, achieving stable operation at approximately 10~FPS per feed, including overhead video capture, real-time bounding-box overlays, and concurrent updates to anomaly score graphs.

However, the current system is designed specifically for person-level anomalies in controlled indoor settings. Anomalies that do not involve people, such as vehicle intrusion, or open-set anomaly types not covered by the four fixed prompts, would require complementary detection modules in a production deployment. The fixed-prompt design was chosen deliberately for zero-training deployability and interpretability: operators can inspect and modify the prompt set without machine-learning expertise. Learnable prompt expansion, including LLM-generated candidate prompts, is deferred to future work.

\section{Conclusion}
\label{sec:conclusion}

This paper presented a two-stage anomaly detection framework that couples YOLO~v11n-pose with CLIP~ViT-B/32, replacing multi-component pipelines that require optical flow and density estimation. The key finding is that direct text-image similarity via CLIP can serve as an effective scoring mechanism, achieving a $3.36\times$ throughput gain (51.39~FPS) while maintaining comparable AUROC on established benchmarks and improving performance on the targeted indoor dataset (84.13\%). Deployment on live CCTV feeds at Chulalongkorn~University confirmed stable operation at over 10~FPS per feed.

The current design has limitations. The fixed set of four text prompts constrains the detectable anomaly categories, as reflected in the lower ShanghaiTech~Campus AUROC (70.26\%) where anomalies such as cycling and vehicle intrusion fall outside prompt coverage. Additionally, the YOLO detector is fine-tuned on indoor scenes, which may reduce generalization to markedly different environments.

Future work will address these limitations by using learnable prompt expansion, inspired by VadCLIP~\cite{b9}, to broaden anomaly coverage; illumination-robust preprocessing for low-light environments; and developing a unified end-to-end model suitable for deployment on resource-constrained edge devices.

\section*{Acknowledgment}

The authors gratefully acknowledge Tatsawal Chantha\-rachana, Siwanont Tantaya\-pirom, and Tanakrit Sutthi\-pattanakit from the Department of Electrical Engineering, Faculty of Engineering, Chulalongkorn~University, for their contributions to the initial development of the YOLO-based detection and CLIP-based scoring components that form the experimental foundation of this work. This work was supported by the Thailand Science Research and Innovation Fund, Chulalongkorn University (IND\_FF\_69\_105\_2100\_018), and by the Video Technology Research Group (VTRG), Department of Electrical Engineering, Faculty of Engineering, Chulalongkorn~University. The first author acknowledges the Graduate Scholarship Program for ASEAN or Non-ASEAN Countries.

\end{document}